\documentclass[11pt,a4paper]{article}
\usepackage{authblk}
\usepackage[hyperref]{acl2020}
\usepackage{times}
\usepackage{latexsym}

\usepackage{url}

\usepackage{amsfonts}  
\usepackage{amssymb}
\usepackage{epstopdf}
\usepackage{enumitem}
\usepackage{mathtools}
\usepackage{bbm}
\usepackage{url}
\usepackage{wrapfig}
\usepackage{appendix}
\usepackage{cases}
\usepackage{enumitem}
\usepackage{flexisym}

\usepackage{helvet}  
\usepackage{courier}  
\usepackage{url}  
\usepackage{graphicx}  
\usepackage[hang,flushmargin]{footmisc}
\usepackage{algorithm}  
\usepackage{algpseudocode}  
\usepackage{amsmath}
\usepackage{multicol}  
\usepackage{multirow} 
\usepackage{cleveref}
\crefformat{section}{\S#2#1#3} 
\crefformat{subsection}{\S#2#1#3}
\crefformat{subsubsection}{\S#2#1#3}

\usepackage{array}
\newcolumntype{L}[1]{>{\raggedright\let\newline\\\arraybackslash\hspace{0pt}}m{#1}}
\newcolumntype{C}[1]{>{\centering\let\newline\\\arraybackslash}m{#1}}
\newcolumntype{R}[1]{>{\raggedleft\let\newline\\\arraybackslash\hspace{0pt}}m{#1}}

\aclfinalcopy 

\title{Stylistic Dialogue Generation via \\Information-Guided Reinforcement Learning Strategy}

\author[1]{Yixuan Su}
\author[2]{Deng Cai}
\author[3]{Yan Wang}
\author[1]{Simon Baker}
\author[1]{Anna Korhonen}
\author[1]{Nigel Collier}
\author[3]{Xiaojiang Liu}
\affil[1]{University of Cambridge}
\affil[2]{The Chinese University of Hong Kong}
\affil[3]{Tencent AI Lab}
\affil[ ]{{\{ys484,sb895,alk23,nhc30\}@cam.ac.uk, thisisjcykcd@gmail.com}}
\affil[ ]{{\{brandenwang,kieranliu\}@tencent.com}}

\date{}

\begin{document}

\maketitle
\begin{abstract}
Stylistic response generation is crucial for building an engaging dialogue system for industrial use. While it has attracted much research interest, existing methods often generate stylistic responses at the cost of the content quality (relevance and fluency). To enable better balance between the content quality and the style, we introduce a new training strategy, know as \textit{\underline{\textbf{I}}nformation-\underline{\textbf{G}}uided \underline{\textbf{R}}einforcement \underline{\textbf{L}}earning} (IG-RL).  In IG-RL, a training model is encouraged to explore stylistic expressions while being constrained to maintain its content quality. This is achieved by adopting reinforcement learning strategy with statistical style information guidance for quality-preserving explorations. Experiments on two datasets show that the proposed approach outperforms several strong baselines in terms of the overall response performance.
\end{abstract}
\section{Introduction}
Most early research on dialogue response generation focused on generating grammatically correct and contextually relevant responses \cite{DBLP:conf/emnlp/RitterCD11,DBLP:journals/sigkdd/ChenLYT17,Martinovsky_2003.theerror}. While good performance has been achieved \cite{DBLP:conf/emnlp/WenGMRSUVY16,DBLP:conf/icassp/WangZTLL16}, syntactically coherent responses alone do not guarantee an engaging and attractive chatbot. In practice, from an industrial point of view, we found that if a chatbot could possess certain language style that is consistent with his/her basic character  (male, female, optimistic, humorous), the users' satisfaction and average rounds of interaction can be notably improved \cite{DBLP:conf/acl/SongZLXH19}. 

While the definition of language style can be specified in different contexts \cite{article_Roberts,bell_1984,bell1997towards,article_Niederhoffer,traugott_1975}, our work refers to language style as any characteristic style of expression, from a purely computational standpoint. For example, gender preference can be regarded as one kind of language style. Considering a conversation context ``\textit{Let's go out of town to relax this weekend!}", it is good for a chatbot with male preference to respond like ``\textit{That's great bro. I will go with my buddies together!}" and with female preference to respond like ``\textit{That's so sweet of you. I will bring my besties!}".  
Besides gender preference, our work is also in line with previous work on dialogue generation with emotion \cite{DBLP:conf/acl/WangZ18,DBLP:conf/aaai/Zhong0M19}; response attitude \cite{DBLP:journals/tacl/NiuB18}, and speaker personality \cite{DBLP:conf/acl/LiGBSGD16}. 

The majority of the existing approaches for stylistic response generation \cite{DBLP:conf/naacl/HuangZTD18,DBLP:conf/aaai/ZhouHZZL18,DBLP:conf/acl/LiGBSGD16,DBLP:conf/acl/WangZ18,DBLP:conf/aaai/Zhong0M19,song-etal-2019-generating} take the style information as an additional input to the generation model and maximize the probability of generating a response given the input query. However, these methods require large parallel corpora (consisting of conversation pairs with specified styles) and often tend to output dull and generic responses \cite{DBLP:conf/naacl/LiGBGD16}. As an alternative, reinforcement learning (RL) can provide a more efficient way to optimize the style expressions contained in the generated responses \cite{DBLP:journals/tacl/NiuB18}. Typically, a style-classifier is adopted as a reward agent to evaluate the style score of generated responses and the generation model is then optimized to generate responses with higher scores.

\begin{table}[tb]
	\setlength{\abovecaptionskip}{3pt}
	\setlength{\belowdisplayskip}{3pt}
	\renewcommand\arraystretch{1.1}
	\centering  
	\small
	\scalebox{0.95}{
	\begin{tabular}{|l|}
		\hline
		{\bf Input Query}: What did you do in the morning?\\ \hline  
		{\bf Gender Style}: Female\\
		\hline
		{\bf Vanilla Seq2seq}: In my old ways.\\
		{\bf Memory Networks}: My husband is very handsome.\\
		{\bf RL}: I went to the school. I like 
		him. I like him.\\ \hline
		{\bf Desired Response} : I had my breakfast with my boyfriend.\\ 
		\hline
	\end{tabular}
	}
	\caption{Examples of response with female gender style}
	\label{tb:1}
\end{table}

However, the RL framework could overemphasize the expression of style at the cost of response quality: because during the RL process, the generation model could learn to fool the style-classier using simple stylistic patterns. We show some examples from our preliminary experiments in Table~\ref{tb:1}. As observed, the RL-based approach first generates generic-style text and then appends a simple phrase \emph{``I like him''} to express a female style, as this phrase receives a high score from the female style classifier. Such tricks bring seemingly high style score but significantly harm the content quality (relevance and fluency). A satisfactory stylistic response should express the desired style on the premise of maintaining a high response quality, as with the last row of Table~\ref{tb:1}.


To address this, we propose a new information-guided reinforcement learning (IG-RL) strategy to better balance the trade-off between the stylistic expression and the content quality. 
Our key idea is to restrict the exploration space of the generation model during training, preventing it from collapsing to some trivial solutions.
Specifically, we separate the vocabulary into two sets, \textit{stylistic words} and \textit{neutral words}, according to the point-wise mutual information (PMI) \cite{church1990word} between words and styles. At the training stage, given the reference response, the model is constrained to maintain the tokens at the positions of neutral words. On the other hand, at the positions of stylistic words, the model is allowed to freely explore the entire vocabulary space to search for words that maximize the reward of the style-classifier and are coherent to the surrounding context. In this way, the generation model learns to generate possible stylistic expressions while maintaining a high response quality. \footnote{Please note that the PMI is required during training only. During inference, it generates stylistic responses without any external signals.}

To facilitate future research in this area, we introduce a new large-scale gender-specific dialogue dataset. Experimental results on this new dataset and another public benchmark dataset demonstrate that the proposed approach fosters dialogue responses that are both stylistic and high in quality. It outperforms standard RL models and other strong baselines in terms of the overall response quality.

In summary, the contributions of this work are: (i) A novel training strategy to train a model to generate stylistic responses under the reinforcement learning paradigm. This strategy can properly balance the trade-off between the style expression and the content quality via an information-guided learning process. Human evaluation shows that the proposed approach can generate responses with both high content quality and desired styles. It significantly outperforms existing methods. (ii) A new gender-specific dialogue dataset which contains over 4.5 million query-response pairs. To our best knowledge, this dataset is the first one focusing on gender-specific dialogue generation and can greatly facilitate further work in this area.

\section{Related Work}
Stylistic dialogue response generation has been an active research area in recent years. \citet{DBLP:conf/acl/LiGBSGD16} proposed a model that represents personality as embeddings and incorporate it into a decoder of a seq2seq model. \citet{DBLP:conf/naacl/HuangZTD18} appended the emotion embeddings into decoder states to generate responses with desired emotions. \citet{DBLP:conf/aaai/Zhong0M19} proposed to embed VAD information into words to control the generation of emotional responses. \citet{DBLP:conf/aaai/ZhouHZZL18} used emotion category embedding, internal emotion state and external emotion memory for emotional dialogue generation. However, explicitly incorporating style information into the model configuration may significantly bias the generation process and cause a drastic drop in the response quality.

For RL-based methods, \citet{DBLP:journals/tacl/NiuB18} train an attitude classifier as the reward agent to guide the learning process. However, due to the nature of unconstrained sampling, the dialogue agent could learn a simple behaviour that expresses the desired style. As a result, little knowledge is actually learned by the model during the training stage which further undermines the quality of the generated responses.

It should be noted that the stylistic dialogue generation is different from the task of text style transfer. Text style transfer aims to rewrite the input sentences such that they possess certain styles, while rigorously preserving their semantic content \cite{DBLP:journals/corr/abs-1901-11333}. On the other hand, stylistic dialogue generation does not aim to preserve the semantic meaning of the input sentences. Instead, it aims to generate responses that are adequate to the input query, while expressing pre-specified styles.

\section{Background}
\label{sec:background}
RL-based systems \cite{DBLP:journals/tacl/NiuB18} first train a style-classifier on an annotated dataset as a reward agent. In the training stage, the dialogue generation model generates a response and observes a style score from the reward agent. The parameters of the generation model are optimized to maximize the expected style score.

The learning objective is typically defined as:
\begin{equation}
    \nonumber
    L(\theta) = -\mathbb{E}_{{\bf \bar{Y}}\sim p_{\theta}}[r_{s^{\prime}}({\bf \bar{Y}}-b)],
    \label{eq:5}
\end{equation}
where $p_{\theta}$ is the policy (probability distribution) defined by the model parameters $\theta$ and $s^{\prime}$ is the desired style, $r_{s^{\prime}}$ is the score of the desired style provided by the style-classifier. ${\bf \bar{Y}} = (\bar{y}_1, ..., \bar{y}_N)$ is the sampled response and $\bar{y}_t$ is the token sampled at time step $t$. The baseline $b$ is used to reduce the variance in the training process. 
\begin{figure}[tb] 
	\centering    
	\setlength{\abovecaptionskip}{3pt}
	\includegraphics[width=0.37\textwidth]{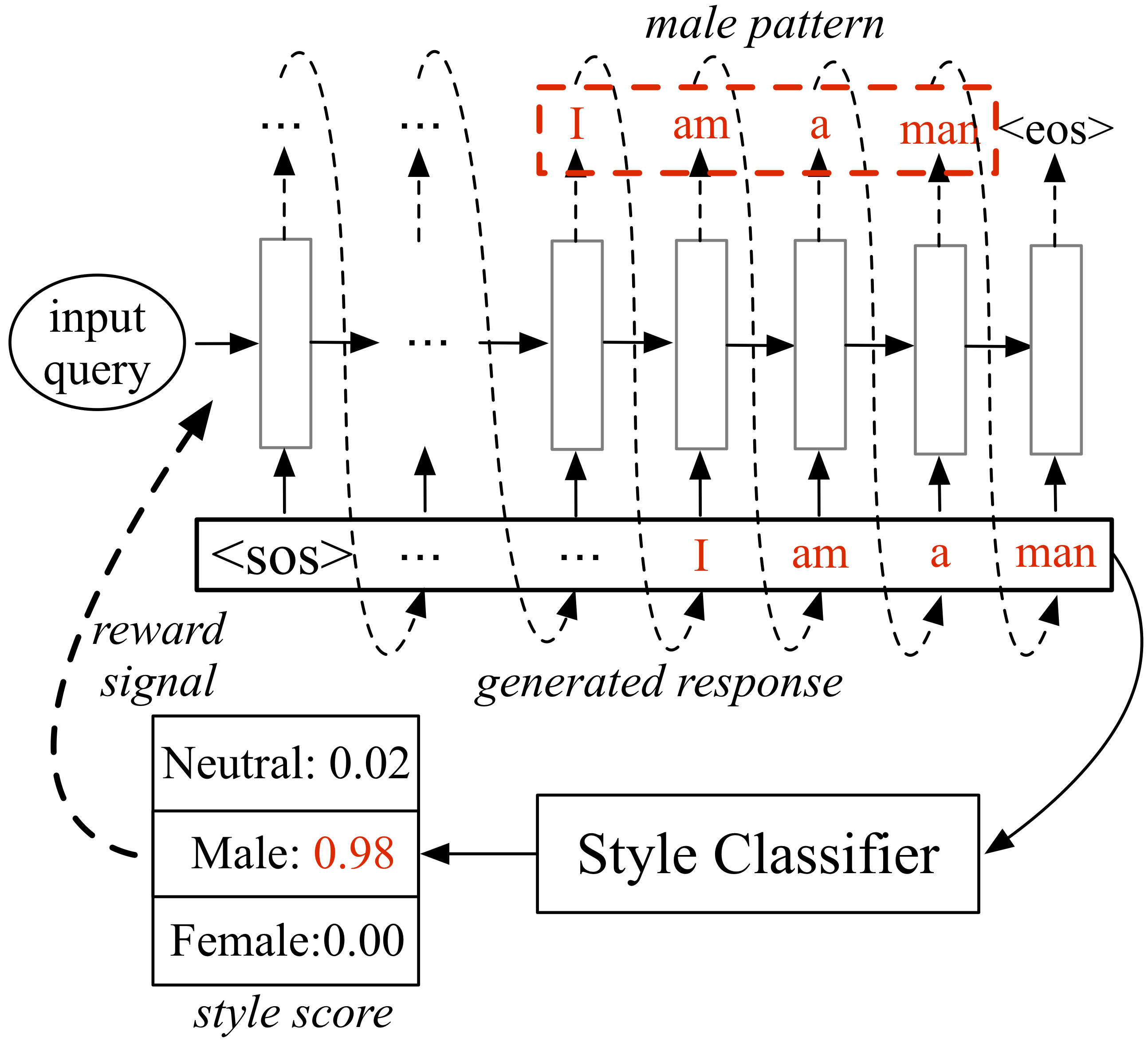}
	\caption{Unconstrained Sampling: by generating a male-style-phrase at the end of response, the generator can easily acquire high scores from the reward agent.}
	\label{fig:unconstrained_sampling}
\end{figure}

Typically, techniques like Monte-Carlo or top-$k$ sampling \cite{DBLP:journals/corr/PaulusXS17} are used to generate response ${\bf \bar{Y}}$ in training. We refer to these approaches as \textit{unconstrained sampling}, since the generated response is solely drawn from the distribution $p_{\theta}$ that is defined by the model parameters. Therefore, the model is allowed to freely explore the entire vocabulary to learn a policy (probability distribution) that optimizes the predefined reward. 

However, conducting efficient exploration is very hard for the unconstrained sampling since the search space is exponentially large, so only frequent patterns which match the reward function are reinforced as shown in Table \ref{tb:1}. Another example is provided in Figure \ref{fig:unconstrained_sampling}. When learning to generate \textit{male} responses, the model learns a simple mechanism that generates a typical male-style-phrase \textit{``I am a man''} at the end of the response to \textit{``cheat''} the reward agent and thus acquire a high male score. Obviously, the learned policy is not ideal since little knowledge other than the simple behaviour is actually learned by the model, and the generated responses can hardly satisfy the users. 

\begin{figure*}[htb] 
	\centering   
	\setlength{\abovecaptionskip}{3pt}
	\includegraphics[width=0.74\textwidth]{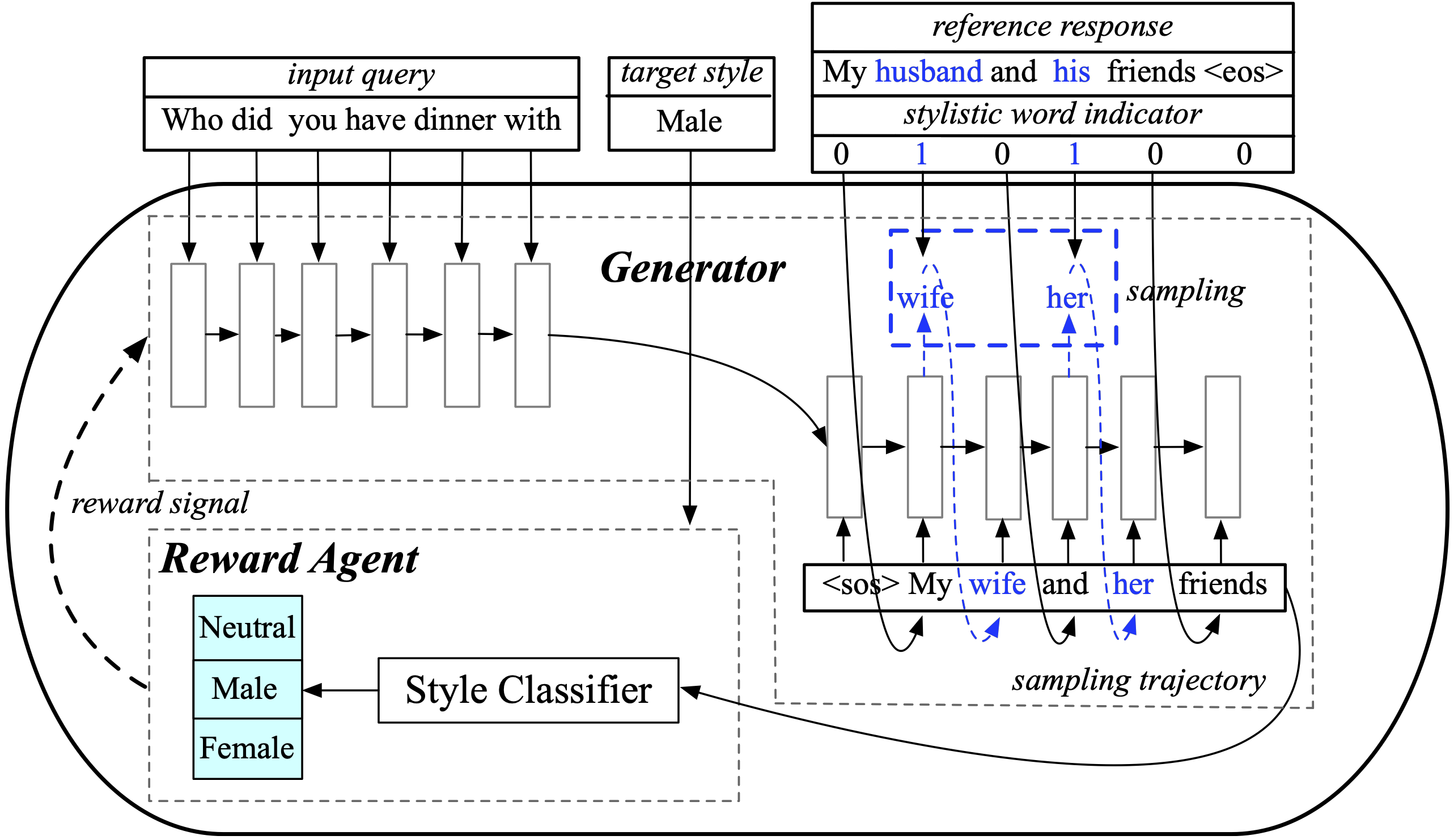}
	\caption{Framework Overview of Information-Guided Reinforcement Learning}
	\label{fig:overview}
\end{figure*}

\section{Information-Guided Reinforcement Learning}
In stylistic dialogue generation task, the training data can be formulated as $(\mathcal{X}, \mathcal{Y}, \mathcal{S})$, where $\mathcal{X}$ is the set of input queries, $\mathcal{Y}$ is the set of responses and $\mathcal{S}$ is the set of all possible styles. Each data instance follows the format of $({\bf X}, {\bf Y}, s)$, where ${\bf X} = (x_1, ..., x_T)$ is the input query, ${\bf Y} = (y_1, ..., y_N)$ is the reference response and $s\in S$ is the style of the reference response ${\bf Y}$. 

To address the problem of unconstrained sampling introduced in \cref{sec:background}, we propose a new training strategy which uses PMI information to guide the training of the generation model under the reinforcement learning framework. 
As illustrated in Figure \ref{fig:overview}, during \textbf{training}, 
stylistic and styleless words are first identified according to the PMI information. Then the model is learned to generate words same with the reference response (``My'', ``and'', ``friends'' in Figure \ref{fig:overview}) at the positions of styleless words, and set free to explore stylistic expressions (``wife'', ``her'' in Figure \ref{fig:overview}) to maximize the expected style score at the positions of stylistic words. Finally, the model parameters are updated via the REINFORCE algorithm \cite{DBLP:conf/nips/SuttonMSM99}. During \textbf{inference}, the model directly generates stylistic responses without any external signals. We denote the proposed training strategy as \textbf{I}nformation-\textbf{G}uided \textbf{R}einforcement \textbf{L}earning (IG-RL) since its training policy is guided by some external information other than sampling in the entire action space (the entire vocabulary set) in an unconstrained manner.



\subsection{Stylistic Words Indication}
\label{sec:PMI}
To indicate whether a token $x$ is stylistic or not given the style $s$, we use the pointwise mutual information (PMI) \cite{church1990word}  which is defined as
\begin{equation}
    \nonumber
    \textup{PMI}(x;s) = \log\frac{p(x,s)}{p(x)p(s)},
    \label{eq:1}
\end{equation}
where $p(x, s)$ is the frequency that the word $x$ appears in a response with style $s$ in the training corpus. We define a word $x$ is stylistic given the style $s$ if $\textup{PMI}(x,s)\geq t_s$. In the experiments, we empirically set $t_s = \frac{3}{4}\times\max_{v\in\mathcal{V}}\textup{PMI}(v; s)$, where $\mathcal{V}$ is the whole vocabulary set.




\subsection{Constrained Sampling}
During the RL training stage, we impose dynamic constraints on the sampled response which is then propagated to the pre-trained classifier to acquire a reward signal. Given a reference response ${\bf Y} = (y_1, ..., y_N)$, PMI between the tokens and the styles are adopted to determine which tokens are stylistic. At sampling step $t$, if $y_t$ is neutral (styleless), then the model is constrained and only permitted to sample $y_t$. 
Otherwise, the model will be permitted to freely sample a new token from the entire vocabulary set. 

The neutral words in the sampled response construct a neutral training skeleton that is closely related to the query. Based on this training skeleton, the model learns to express the desired style by sampling at the positions selected by PMI. An illustration can be found in the right part of Figure \ref{fig:overview}, where the model learns to generate male responses. 
In this example, in reference response ``\textit{My husband and his friends}", ``\textit{husband}" and ``\textit{his}" are denoted as stylistic words. By masking these stylistic words, a neutral training skeleton ``\textit{My \underline{ } and \underline{ } friends}" is constructed. The model is only permitted to sampling new words at the masked positions, and the desired response is ``\textit{My wife and her friends}" which has high content quality and expresses the desired style (male).

\renewcommand{\algorithmicrequire}{\textbf{Input:}}  
\renewcommand{\algorithmicensure}{\textbf{Output:}} 
\begin{algorithm}[tb]  
    \fontsize{11pt}{11pt}\selectfont
	\caption{Constrained Sampling}  
	\begin{algorithmic}[1] 
		\Require 
		Input query ${\bf X} = (x_1, ..., x_T)$; 
		Reference response ${\bf Y} = (y_1, ..., y_N)$;
        Reference response style $s$;
		\Ensure Constrained sampling trajectory ${\bf \bar{Y}}$
		\State ${\bf \bar{Y}} \gets (\langle\textup{Start of Sentence}\rangle)$
		\For{$i = 1$ to $N$}
		\If {$\textup{PMI}(y_i; s) \geq t_s$} 
		\State Sample $\bar{y}_i\sim p_{\theta}(\cdot|{\bf \bar{Y}}; {\bf X})$ 
		\Else
		\State $\bar{y}_i\gets y_i$
		\EndIf
		\State ${\bf \bar{Y}} \gets {\bf \bar{Y}}\cup \bar{y}_i$
		\EndFor  
	\end{algorithmic}  
	\label{alg:sample}
\end{algorithm}


The detailed description of the proposed approach is presented in Algorithm \ref{alg:sample}, where $t_s$ is the style-specific threshold as described in \cref{sec:PMI}.




\subsection{Optimization}
Given a sampling trajectory ${\bf \bar{Y}}$, based on the REINFORCE algorithm, the learning objective is described as 
\begin{equation}
\nonumber
\begin{split}
 &L_{\textup{RL}}(\theta)   \approx -(r_{s^{\prime}}({\bf \bar{Y}}) - b)\log p_{\theta}({\bf \bar{Y}}) \\
    & = -(r_{s^{\prime}}({\bf \bar{Y}}) - b)\sum_{\bar{y}_i\in{\bf \bar{Y}}}\log p_{\theta}(\bar{y}_i|\bar{y}_1,...,\bar{y}_{i-1};{\bf X}),
\end{split}
\label{eq:9}
\end{equation}
where $r_{s^{\prime}}$ is the score of the desired style $s^{\prime}$, The baseline $b$ is used to reduce the variance during the training process and ${\bf X}$ is the input query.

To optimize this objective, ${\bf \bar{Y}}$ should satisfy both the reward agent and the conditional language model. Since the sampling process is dependent on a neutral skeleton, the model has to learn to sample words that not only express the desired style but are also compatible with its context (the surrounding neutral skeleton).

To stabilize the training process, we incorporate a standard Maximum Likelihood Estimation (MLE) objective. Given the input query ${\bf X}$ and the reference response ${\bf Y}$, the objective is defined as
\begin{equation}
    \nonumber
    L_{\textup{MLE}}(\theta) = -\sum_{i=1}^{N}\log p_{\theta}(y_i|y_1, ..., y_{i-1}; {\bf X}).
    \label{eq:10}
\end{equation}


In addition, the MLE objective tends to train a model that is overfit to the training set \cite{DBLP:journals/corr/PereyraTCKH17}, therefore the model is less willingly to explore other possibilities during the RL training process. To mitigate this side effect, we use label smoothing \cite{DBLP:conf/cvpr/SzegedyVISW16} as an auxiliary regularization. 
Instead of using a uniform distribution over all words in the vocabulary as target, we introduce a new form of target distribution. In which case, we use the bigram frequency distribution that acquired from the training corpus as target and the detailed computation is shown as
\begin{equation} 
\nonumber
\begin{split}
    &L_{\textup{smo}}(\theta)  =\\ &-\sum_{i=2}^{N}\sum_{v\in\mathcal{V}}f(y_{i-1}, v)\log p_{\theta}(v|y_1,..., y_{i-1}; {\bf X}), \\
    &f(y_{i-1}, v)  = \frac{\#(y_{i-1}, v)}{\sum_{v^{\star}\in\mathcal{V}}\#(y_{i-1}, v^{\star})},
\end{split}
\label{eq:11}
\end{equation}
where $\#(y_{i-1}, v)$ is the bigram count of token $y_{i-1}$ and $v$ in the training corpus. 




The final learning objective is defined as
\begin{equation}
\nonumber
    L_{\textup{hybrid}}(\theta) = L_{\textup{MLE}}(\theta) + \alpha L_{\textup{smo}}(\theta) + \beta L_{\textup{RL}}(\theta),
    \label{eq:12}
\end{equation}
where $\alpha$ and $\beta$ are weights of different parts.

\begin{table}[t] 
	\centering    
	\setlength{\abovecaptionskip}{3pt}
	\includegraphics[width=0.43\textwidth]{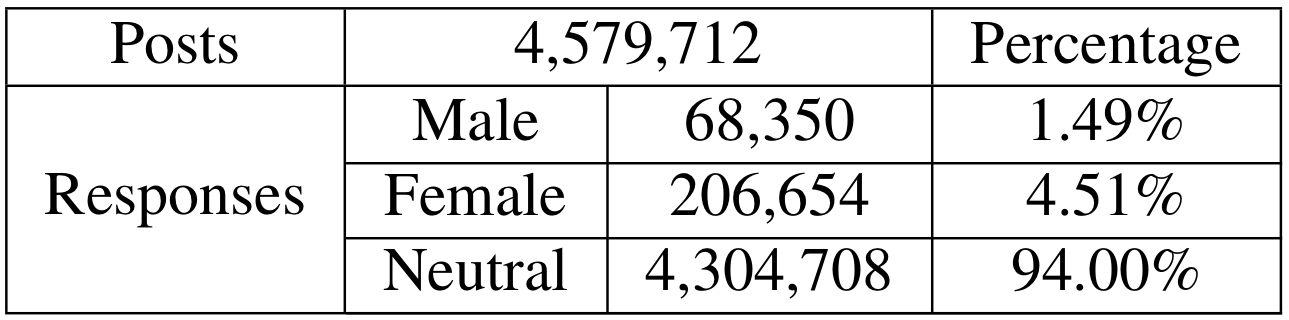}
	\caption{Gender-specific Dialogue Dataset summary}
	\label{tb:gender}
\end{table}

\section{Experiments}
\subsection{Datasets}
To facilitate future research in this area, we constructed a gender-specific dialogue dataset. At the first step, based on the gender information of users, we collected 100,000 query-response pairs whose responses generated by female users from Douban\footnote{https://www.douban.com}. Similarly, we also collected 100,000 query-response pairs from male users. 

Then we hired $6$ professional annotators ($3$ of them are female and others are male) to further verify the collected results. Because there is no rigorous guideline on how to quantify the gender preference expressed in daily life conversation, we let the annotators to judge the results based on their own understanding. The annotators are asked to assign a \textit{female(male)} label to the response if it is very unlikely uttered by a male(female) user. Otherwise, a \textit{neutral} label will be assigned. To ensure a reasonable annotation, each response is annotated by all six annotators, and we only keep the pairs whose response label is agreed by at least five annotators.

From the 200,000 collected pairs, 5,184 responses are annotated as male instances and 10,710 responses are annotated as female instances. To keep data statistic balanced, we randomly select 15,000 neutral instances to build the high-quality gender classification dataset. Then we fine-tuned a Chinese BERT \cite{DBLP:conf/naacl/DevlinCLT19} on the constructed dataset to build a gender-classifier. And the classification accuracy is about 91.7\%. We further use this classifier to automatically annotate the STC-Sefun dataset \cite{DBLP:conf/acl/BiGLS19} to obtain a big gender-specific dialogue dataset\footnote{The released version of the proposed dataset can be found here: https://ai.tencent.com/ailab/nlp/dialogue/\#datasets} whose data statistic is shown in Table \ref{tb:gender}. 

\begin{table*}[tb]
	\setlength{\abovecaptionskip}{3pt}
	\renewcommand\arraystretch{1.1}
	\centering  
    \begin{center}
    \scalebox{0.6}{
    \begin{tabular}{|C{2cm}|c|C{1.8cm}|C{1.8cm}|C{1.8cm}|C{1.8cm}|C{1.8cm}|C{1.8cm}|}
	\hline
	\multirow{ 2}{*}{Style}&\multirow{ 2}{*}{Metric}&\multicolumn{4}{|c|}{Baselines}&\multicolumn{2}{|c|}{Ours}\\
	\cline{3-8}
	&&Seq2seq&Speaker&ECM&Polite-RL&w/o G&IG-RL\\\hline
	\hline
    \multirow{3}{*}{Female}&Quality$\uparrow$&\textbf{3.24}${\dagger}$&1.45&2.22&2.33&2.96${\dagger}$&2.99\\
    &Style Expression$\uparrow$&3.03&3.26&3.19&3.63${\dagger}$&3.60${\dagger}$&\textbf{3.64}\\
    &Ranking$\downarrow$&2.77&3.56&3.20&2.81&2.01&\textbf{1.75}\\
    \hline
    \multirow{3}{*}{Male}&Quality$\uparrow$&\textbf{3.13}${\dagger}$&1.41&1.97&2.31&2.93&3.02\\
    &Style Expression$\uparrow$&2.99&3.56&3.49&\textbf{4.58}${\dagger}$&3.25&4.03\\
    &Ranking$\downarrow$&2.94&3.75&3.42&2.13&2.71&\textbf{1.72}\\
    \hline
    \multirow{5}{*}{Overall}&Quality$\uparrow$&\textbf{3.19}${\dagger}$&1.43&2.10&2.32&2.94&3.01\\
    &Style Expression$\uparrow$&3.01&3.41&3.34&\textbf{4.11}${\dagger}$&3.43&3.84\\
    &Ranking$\downarrow$&2.86&3.66&3.31&2.47&2.36&\textbf{1.73}\\
    \cline{2-8}
    &Distinct-1(\%)$\uparrow$&19.73&15.95&13.77&13.22&20.23&\textbf{24.92}\\
    &Distinct-2(\%)$\uparrow$&67.81&61.05&59.72&55.02&70.44&\textbf{74.83}\\
    \hline
	\end{tabular}}
	\end{center}
	\caption{Evaluation Results on Gender-Specific Dialogue Generation: $\uparrow$ means the higher the better and $\downarrow$ means the lower the better, bold font denotes the best scores for each metric. Sign tests on evaluation scores show that the proposed model significantly outperforms other models with p-value $<$ 0.05 with the only exception marked by ${\dagger}$.}
	\label{tb:gen_table}
\end{table*}

\begin{table*}[tb]
	\setlength{\abovecaptionskip}{3pt}
	\renewcommand\arraystretch{1.1}
	\centering  
    \begin{center}
    \scalebox{0.6}{
	\begin{tabular}{|C{2cm}|c|C{1.8cm}|C{1.8cm}|C{1.8cm}|C{1.8cm}|C{1.8cm}|C{1.8cm}|}
	\hline
	\multirow{2}{*}{Style}&\multirow{ 2}{*}{Metric}&\multicolumn{4}{|c|}{Baselines}&\multicolumn{2}{|c|}{Ours}\\
	\cline{3-8}
	&&Seq2seq&Speaker&ECM&Polite-RL&w/o G&IG-RL\\\hline
	\hline
    \multirow{3}{*}{Like}&Quality$\uparrow$&\textbf{3.13}${\dagger}$&2.41&2.38&2.48&3.01${\dagger}$&2.96\\
    &Style Expression$\uparrow$&2.90&3.72${\dagger}$&3.78${\dagger}$&\textbf{4.42}${\dagger}$&3.27&3.66\\
    &Ranking$\downarrow$&3.35&3.14&3.24&2.51&2.59&\textbf{2.14}\\
    \hline
    \multirow{3}{*}{Disgust}&Quality$\uparrow$&\textbf{3.03}${\dagger}$&2.24&2.15&2.29&2.95${\dagger}$&2.65\\
    &Style Expression$\uparrow$&2.82&3.58&3.81&\textbf{4.59}&2.94&3.72\\
    &Ranking$\downarrow$&3.26&3.49&3.39&2.33&2.68&\textbf{2.24}\\
    \hline
    \multirow{3}{*}{Happiness}&Quality$\uparrow$&3.08&2.51&2.39&2.69&2.93&\textbf{3.27}\\
    &Style Expression$\uparrow$&3.45&4.78${\dagger}$&4.77${\dagger}$&4.73${\dagger}$&4.34&\textbf{4.79}\\
    &Ranking$\downarrow$&3.49&2.64&2.89&2.49&2.36&\textbf{1.56}\\
    \hline
    \multirow{3}{*}{Anger}&Quality$\uparrow$&\textbf{3.24}${\dagger}$&2.43&2.15&2.06&2.99${\dagger}$&2.84\\
    &Style Expression$\uparrow$&2.46&3.82${\dagger}$&4.02${\dagger}$&\textbf{4.15}${\dagger}$&2.78&3.87\\
    &Ranking$\downarrow$&3.39&2.92&3.10&3.14&2.71&\textbf{1.86}\\
    \hline
    \multirow{3}{*}{Sadness}&Quality$\uparrow$&\textbf{3.00}${\dagger}$&2.10&2.04&2.24&2.77${\dagger}$&2.71\\
    &Style Expression$\uparrow$&2.49&3.99${\dagger}$&4.08${\dagger}$&\textbf{4.45}${\dagger}$&2.78&3.93\\
    &Ranking$\downarrow$&3.67&3.06&3.26&2.41&3.01&\textbf{1.98}\\
    \hline
    \multirow{3}{*}{Overall}&Quality$\uparrow$&\textbf{3.09}${\dagger}$&2.34&2.22&2.35&2.93${\dagger}$&2.89\\
    &Style Expression$\uparrow$&2.82&3.98${\dagger}$&4.09${\dagger}$&\textbf{4.47}${\dagger}$&3.22&3.99\\
    &Ranking$\downarrow$&3.43&3.05&3.18&2.56&2.67&\textbf{1.96}\\
    \cline{2-8}
    &Distinct-1(\%)$\uparrow$&17.41&15.69&13.65&11.50&19.59${\dagger}$&\textbf{20.46}\\
    &Distinct-2(\%)$\uparrow$&65.07&58.39&52.19&51.12&67.34&\textbf{73.37}\\
    \hline
	\end{tabular}}
	\end{center}
	\caption{Evaluation Results on Emotion-Specific Dialogue Generation}
	\label{tb:emo_table}
\end{table*}

For a comprehensive evaluation, in addition to the proposed gender-specific dialogue dataset, we also conduct experiments on a public emotional dialogue dataset \cite{DBLP:conf/aaai/ZhouHZZL18}. 

\subsection{Implementation Details}
The proposed model is implemented using Pytorch \cite{paszke2017automatic}. We use two-layer LSTMs with 500 hidden units to construct the encoder and decoder of the generation model. The word embedding size is set to 300 and it is randomly initialized. The vocabulary size is limited to 15,000.
 
We use Adam \cite{DBLP:journals/corr/KingmaB14} to optimize our model with a batch size of 64 and learning rate of 1e-3. For all experiments, we first pretrain a seq2seq model with the MLE objective for 3 epoches on the training set. Then the learned parameters are used to initialize the policy networks. We set the reference reward $b$ and $\alpha$, $\beta$ in the learning objective as 0.3, 0.2, and 0.25 respectively. Similar to recent works \cite{DBLP:conf/acl/LewisDF18,DBLP:conf/acl/QinGBLGDCG19}, we use top-$k$ sampling during the inference stage with $k$ set to 20.

\subsection{Compared Models}
We compare the proposed approach with several representative and competitive baselines.
\subsubsection{Baselines}
\paragraph{Seq2seq:} Standard sequence-to-sequence model with attention mechanism \cite{DBLP:conf/emnlp/LuongPM15}.

\paragraph{Speaker:} Model proposed by \citet{DBLP:conf/acl/LiGBSGD16} which incorporates distributed style embeddings into the structure of decoding cell to control the generation process.

\paragraph{ECM:} Model proposed by \citet{DBLP:conf/aaai/ZhouHZZL18} which adopts internal and external memory modules to control the generation of stylistic expressions in the generated responses.


\paragraph{Polite-RL:} Approach proposed by \citet{DBLP:journals/tacl/NiuB18} which leverages RL to teach the model to generate stylistic responses. In our experiments, we also use BERT as the reward agent.

\subsubsection{Ablation Study}
\paragraph{IG-RL:} The full model proposed in this work. For a fair comparison, we construct our generator using the same structure as the one in \citet{DBLP:journals/tacl/NiuB18}.

\paragraph{w/o G:} In the ablated model, we examine how the guidance provided by PMI knowledge effects the model's performance. To this end, in the RL training stage, instead of using PMI to guide the sampling process, we let the model sample a token with a equal probability of 0.2 or just simply copy the corresponding token in the reference response. 

\subsection{Evaluation Metrics}
The quality of the responses generated by a dialogue system is well known to be difficult to measure automatically \cite{DBLP:journals/corr/abs-1905-04071}; therefore we rely on human evaluation. In our experiments, each response is judged by five independent annotators who are hired from a commercial company. To prevent possible bias from the annotators, all results are randomly shuffled before being evaluated and are evaluated following metrics below.

\paragraph{Quality:} This metric evaluates the content quality of the generated responses. The annotators are asked to give a score within $5$-point scale where $5$ means \textit{perfectly human-like response} (relevant, fluent and informative), and $1$ means \textit{unreadable}. 

\paragraph{Style Expression:} This metric measures how well the generated responses express the desired style. The annotators give a score within $5$-point scale, where $5$ means \textit{very strong style}, $3$ means \textit{neutral or no obvious style} and $1$ means \textit{very conflicted style}. The style conflict means the generated style is conflicted to the desired one (e.g. female to male, positive to negative emotion). 

\paragraph{Ranking:} The annotators are further asked to jointly evaluate the content quality and the style expression of the generated responses from different approaches. Then the annotators give a ranking to each result where top $1$ means the best.



We measure the agreement of our annotators using Fleiss$\textprime$ kappa \cite{fleiss1971mns}: for the gender-specific dialogue generation, the results of \textit{Quality} and \textit{Style Expression} are 0.442 and 0.654, which indicate ``moderate agreement" and ``substantial agreement" respectively. As for emotional dialogue generation, the results are 0.432 and 0.628, which indicate ``moderate agreement" and ``substantial agreement" respectively.

\subsection{Main Results}
The evaluation results are shown in Tables \ref{tb:gen_table} and \ref{tb:emo_table} in which we also present the averaged evaluation scores among different styles.

From the results, we can see that the IG-RL method achieves top two performances on both the quality metric and the style metric for both datasets. Compared to other methods, it ensures both high quality and desired stylistic expression. For the ranking metric which jointly evaluates both the content quality and the style expression, the proposed approach outperforms all other baselines by a substantial margin. In addition, we also measure the diversity of the generated responses with two automatic metrics: Distinct-1 and Distinct-2 \cite{DBLP:conf/acl/LiGBSGD16}, and the results show that the IG-RL method generates the most diverse responses among all methods.

It can be observed that Polite-RL generally obtains the highest style expression score but gets much lower performance on the quality and the ranking metric comparing to the proposed approach. This confirms our early analysis that the vanilla RL methods may achieve high style intensity at the cost of the content quality.



The performance on individual styles also provides some insights. In the happiness style, the proposed approach achieves the highest scores on all three metrics. The reason is that the happiness style is similar to the neutral style and we have relatively sufficient data. A similar phenomenon can also be found in the female style response. Therefore, we can conclude that for those styles with sufficient data, the proposed IG-RL can achieve high performance on both of the quality and style aspects. On the other hand, when stylistic data is limited, it also maintains a well balance between the response quality and the style expression.


\begin{table*}[tb] 
	\centering    
	\setlength{\abovecaptionskip}{3pt}	\includegraphics[width=0.83\textwidth]{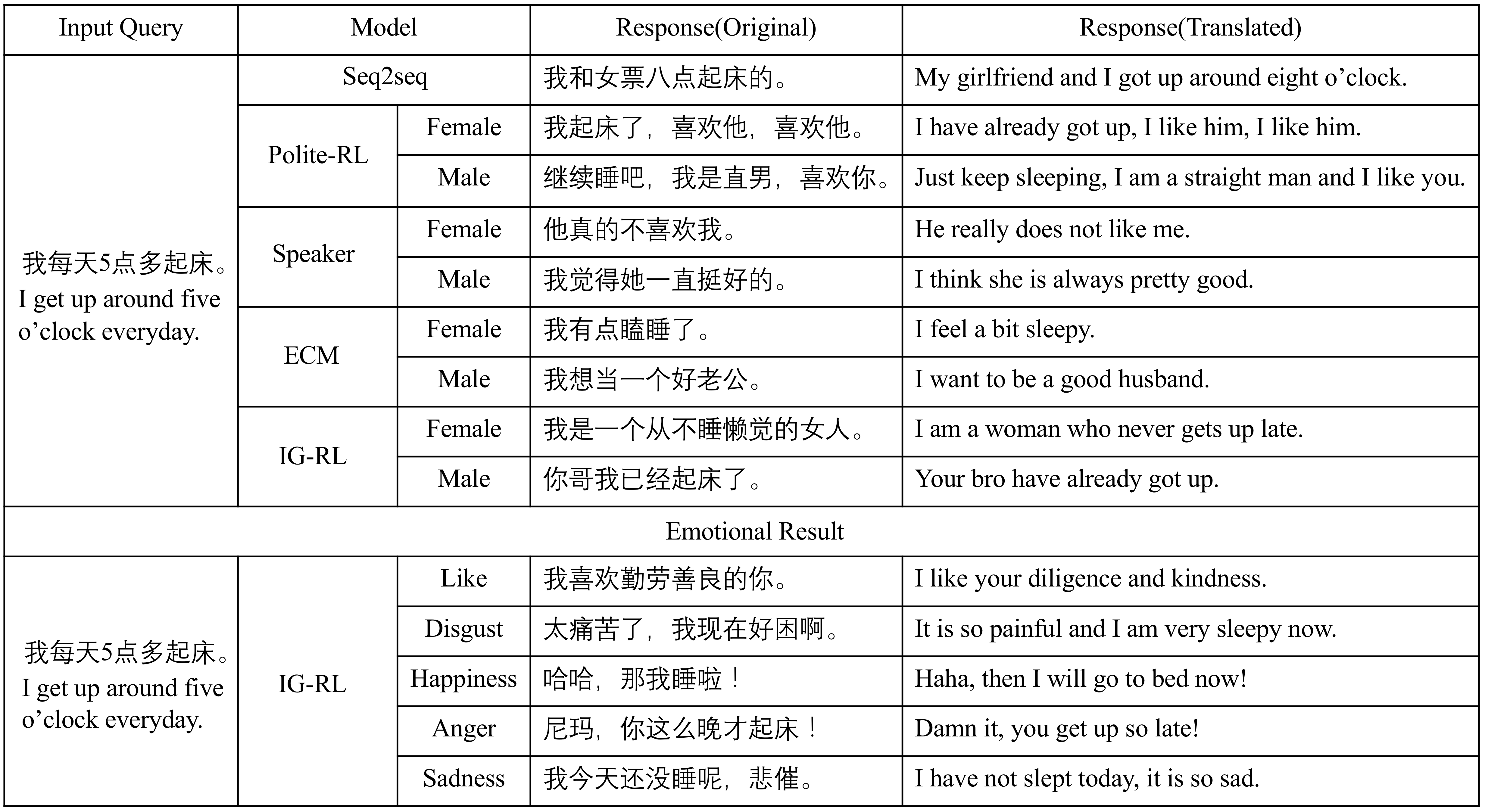}
	\caption{Sample responses generated by different approaches, the input query does not appear in the training set.}
	\label{fig:case_study}
\end{table*}

\subsection{Further Analysis}
Here, we present further discussion and empirical analysis of the proposed method.
\subsubsection{Style Acceptance} 
A fundamental requirement of a stylistic dialogue system is to generate responses that are not conflicted with respect to the desired style. For instance, generating male style responses is not acceptable for a female chatbot; likewise, for a positive emotional chatbot (e.g. like, happiness), generating negative responses (e.g. disgust, sadness and anger) is not acceptable.
\begin{figure}[tb] 
	\centering    
	\setlength{\abovecaptionskip}{3pt}
	\includegraphics[width=0.35\textwidth]{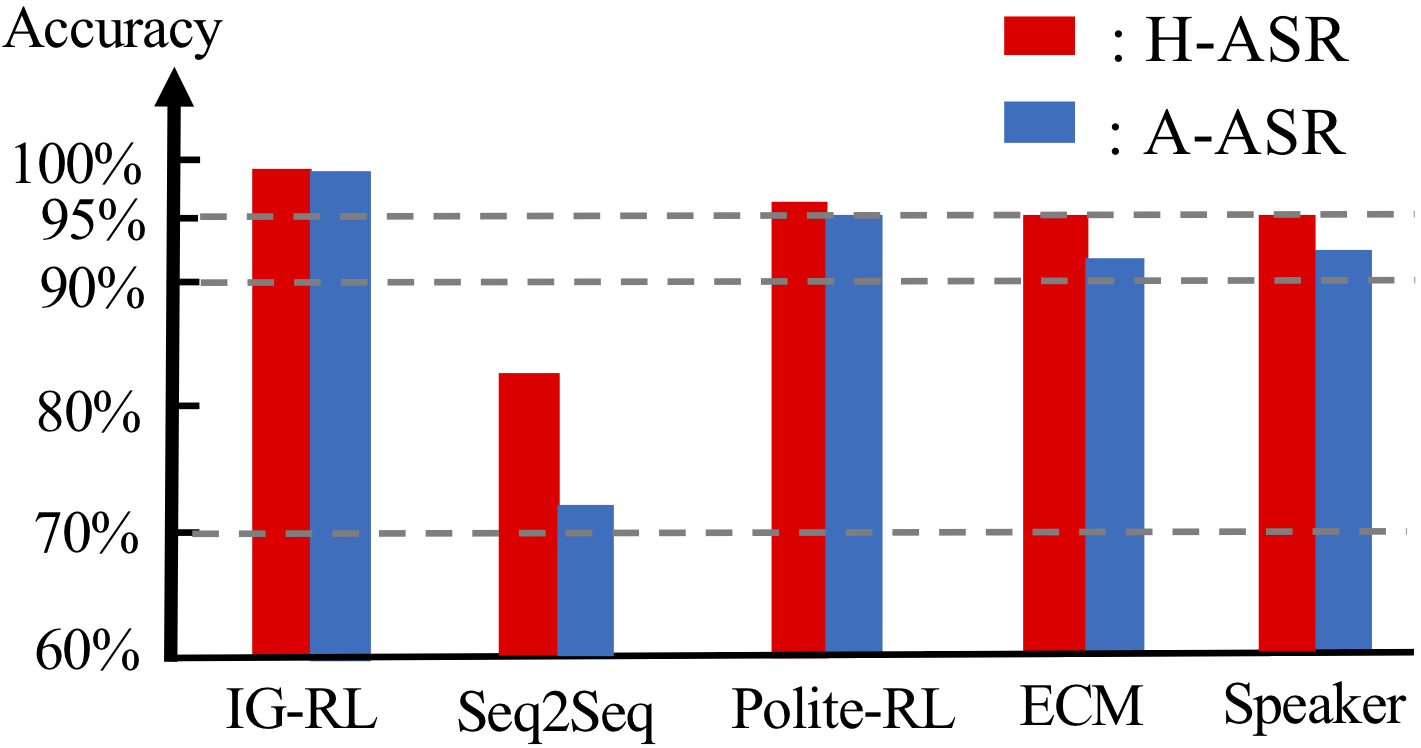}
	\caption{Style Acceptance Evaluation}
	\label{tb:cla_acc}
\end{figure}
To quantitatively evaluate how acceptable a stylistic dialogue system is, we propose two novel metrics: human style acceptance rate (H-SAR) and, automatic style acceptance rate (A-SAR). We compute H-SAR based on the style expression scores in human evaluation. It is defined as the ratio of generated results whose style expression score is greater or equal to $3$. As for A-SAR, we use the pre-trained style-classifier to compute the ratio of the generated responses that display a style which is not conflicted to the desired one. 

\begin{figure}[tb] 
	\centering    
	\setlength{\abovecaptionskip}{3pt}
	\includegraphics[width=0.35\textwidth]{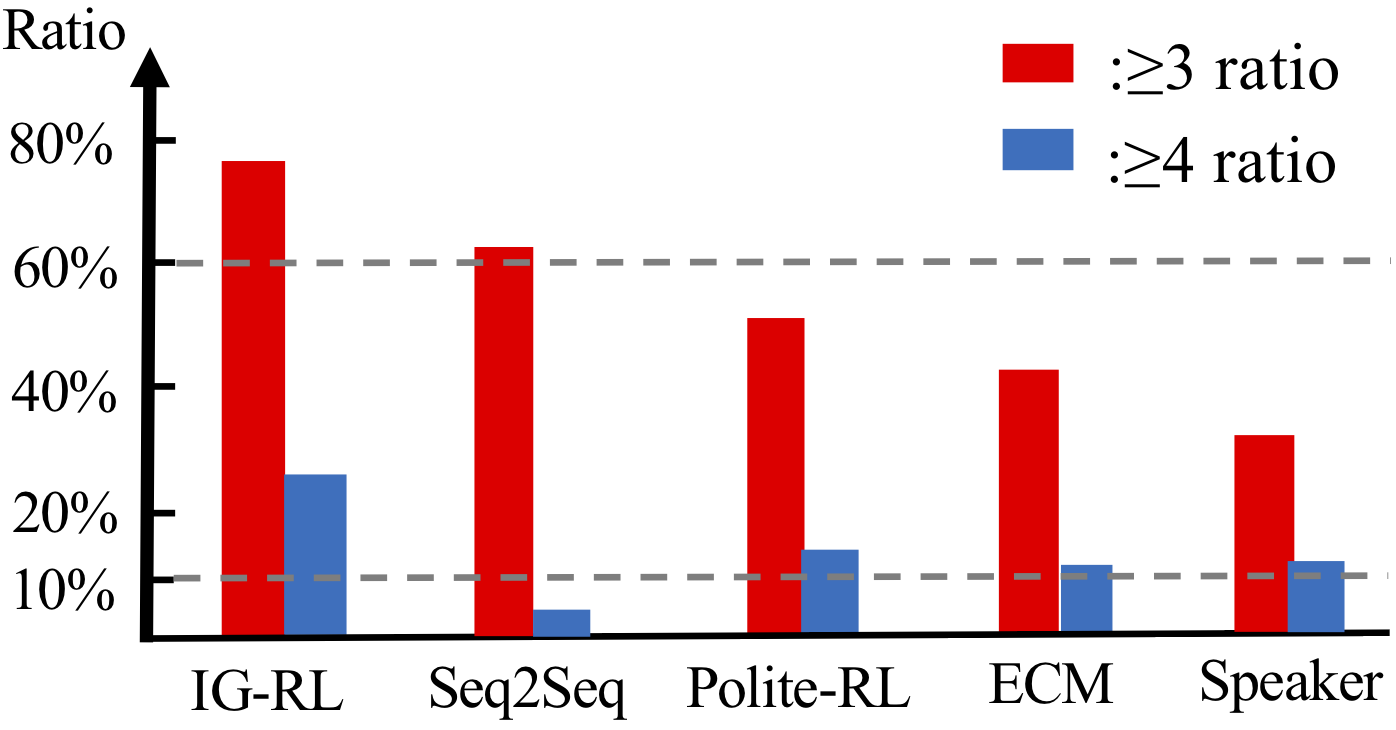}
	\caption{Balance between Quality and Style: The \textit{$\geq3$-ratio} means the ratio of responses whose both scores are greater or equal to $3$; \textit{$\geq4$-ratio} means the ratio of responses whose both scores are greater or equal to $4$.}
	\label{tb:balance}
\end{figure}

The results are shown in Figure \ref{tb:cla_acc} and we can see that H-SAR and A-SAR are highly correlated. Considering results in Table \ref{tb:gen_table} and \ref{tb:emo_table}, although the proposed approach does not generate responses with the highest style expression score, it is the only system which achieves the best H-SAR and A-SAR performances, suggesting that our system is more robust than others since it makes fewer style conflict mistakes.




\subsubsection{Balance between Quality and Style}
A satisfactory stylistic dialogue system should express the desired style while maintaining the content quality. Based on the human evaluation metric, $3$ is the marginal score of acceptance. So we deem a response as marginally acceptable by actual users when both the quality and style expression scores are greater or equal to $3$. On the other hand, $4$ is the score that well satisfies the users, so responses with both scores greater or equal to $4$ are deemed as satisfying to actual users. 

The ratios of both scores $\geq3$ and $\geq4$ are shown in Figure \ref{tb:balance}, from which we see that our system outperforms all other systems on \textit{$\geq3$-ratio} and \textit{$\geq4$-ratio}. Obviously, the proposed IG-RL best balances the trade-off between the response quality and the style expression and therefore generating the most acceptable and satisfying responses.

\subsubsection{Ablation Study} 
We analyze the effect of removing guidance provided by the PMI signal. Comparing the ablated model (w/o G) with our full model (IG-RL), from results in Tables \ref{tb:gen_table} and \ref{tb:emo_table}, we can observe that the quality score is slightly influenced, but the style expression score drops significantly. This demonstrates that although utilizing reference response helps in maintaining the response quality, the guidance provided by PMI information is indispensable for generating stylistic responses.

\subsubsection{Case Study}
We use an input query that is unseen in both datasets to generate responses with different styles using different systems (example responses presented in Table \ref{fig:case_study}). Due to limited space, we only compare different approaches with respect to genders. As for emotions, we present the results of IG-RL only.

We can see that although the result from Seq2seq approach is relevant but it is conflict to female style. As for other compared methods, the memory networks-based approaches (Speaker, ECM) can generate responses with the desired style but not very relevant to the input query. For Polite-RL approach, it first generates part of the response that relates to the input query and then simply generates a phrase which expresses intense desired style (e.g. ``I like him" in female responses). Given all genders and emotions, only the responses generated by IG-RL generally maintain high content quality and properly express the desired style.


\section{Conclusion}
We have proposed a new training strategy that leverages stylistic information as guidance to conduct a quality-preserving learning process. To facilitate future research, we have also constructed and annotated a new dataset for gender-specific dialogue generation. Our experimental results demonstrate that the proposed IG-RL approach outperforms existing baselines in terms of the overall response performance.


\bibliography{acl2020}
\bibliographystyle{acl_natbib}
\end{document}